\pdfoutput=1

\documentclass[11pt]{article}

\usepackage[]{acl}

\usepackage{times}
\usepackage{latexsym}

\usepackage[T1]{fontenc}

\usepackage[utf8]{inputenc}

\usepackage{microtype}

\usepackage{cite}
\usepackage{amsmath,amssymb,amsfonts}
\usepackage{algorithmic}
\usepackage{graphicx}
\usepackage{textcomp}
\usepackage{xcolor}
\usepackage{subcaption}
\usepackage{amsmath}
\usepackage{microtype}
\usepackage{tabularx}
\usepackage{caption}

\DeclareMathOperator*{\argmax}{arg\,max}

\DeclareMathOperator{\entity}{entity}
\DeclareMathOperator{\cond}{cond}

\usepackage{times}
\usepackage{latexsym}

\newcommand\rot[1]{\rlap{\rotatebox{45}{#1}}}

\title{Flowstorm: Open-Source Platform with Hybrid Dialogue Architecture}

\author {
    Jan Pichl,\textsuperscript{\rm 1}\textsuperscript{\rm 2}
    Petr Marek, \textsuperscript{\rm 1}\textsuperscript{\rm 2}
    Jakub Konr\'ad \textsuperscript{\rm 1}\textsuperscript{\rm 2}
    Petr Lorenc \textsuperscript{\rm 1}\textsuperscript{\rm 2} \\
    \bf Ondřej Kobza\textsuperscript{\rm 1}
    Tomáš Zajíček \textsuperscript{\rm 2}
    Jan Šedivý \textsuperscript{\rm 3}\textsuperscript{\rm 2} \\
    \textsuperscript{\rm 1} Faculty of Electrical Engineering, Czech Technical University, Prague, Czech Republic\\
    \textsuperscript{\rm 2} PromethistAI, Prague, Czech Republic\\
    \textsuperscript{\rm 3} CIIRC, Czech Technical University, Prague, Czech Republic\\
    \texttt{\{pichljan,marekp17,konrajak,lorenpe2,kobzaond\}@fel.cvut.cz,}\\
    \texttt{\{tomas.zajicek,jan.sedivy\}@promethist.ai}
}

\begin{document}
\maketitle
\begin{abstract}
This paper presents a conversational AI platform called Flowstorm. Flowstorm is an open-source SaaS project suitable for creating, running, and analyzing conversational applications. Thanks to the fast and fully automated build process, the dialogues created within the platform can be executed in seconds. Furthermore, we propose a novel dialogue architecture that uses a combination of tree structures with generative models. The tree structures are also used for training NLU models suitable for specific dialogue scenarios. However, the generative models are globally used across applications and extend the functionality of the dialogue trees. Moreover, the platform functionality benefits from out-of-the-box components, such as the one responsible for extracting data from utterances or working with crawled data. Additionally, it can be extended using a custom code directly in the platform. One of the essential features of the platform is the possibility to reuse the created assets across applications. There is a library of prepared assets where each developer can contribute. All of the features are available through a user-friendly visual editor. 
\end{abstract}

\section{Introduction}

With the increasing popularity of conversational systems in various areas, there is an increasing demand for human-like aspects of the systems. The human-like aspects cover not only the content of the conversation but also the conversational style and tone of voice. Moreover, many conversational constructs are applicable to various situations and are therefore suitable for sharing and reusing. We introduce the Flowstorm platform\footnote{\url{https://app.flowstorm.ai}}, where conversational designers can easily design an application using shared assets created by a community, create an application on their own, or use a combination of both approaches. The platform does not require any installation as it can be used as a web app. However, advanced developers can use the open-source code\footnote{\url{https://gitlab.com/promethistai/flowstorm}} to run and modify their own version of the platform.

Our platform works with a novel dialogue architecture we have proposed. It uses a combination of dialogue trees/graphs (scenarios) specific to a domain (or only a part of the domain) and generative models handling the aspects of the conversation that are not domain-specific. The Flowstorm platform targets users with limited coding skills and conversational AI knowledge as well as the experts in the area who can easily extend an out-of-the-box functionality with custom logic. Using the provided tools, documentation\footnote{\url{https://docs.flowstorm.ai/}}, and library of the shared assets, one can create a dialogue application in minutes. The Alquist \citep{2021alquist} socialbot, which was the first place winner in the Alexa Prize Socialbot Grand Challenge 4 \citep{shui2021further},  was created using the Flowstorm platform, proving the platform to be suitable even for such a complex system.

Flowstorm platform supports English, Czech, German and Spanish languages.

\section{Related Work}
There are two categories of related software: open-source frameworks or libraries and online platforms. Rasa \citep{bocklisch2017rasa} and DeepPavlov \citep{burtsev2018deeppavlov} are the most famous representatives of the framework category. They provide the developer with a set of programmatic and declarative interfaces for defining and training natural language understanding (NLU) and dialogue management (DM) models. They also allow creating a pipeline of the components where the developer can create a sequence of models or components for utterance processing. Both libraries require a local installation and, in addition, integrating either of them with the rest of the conversational application requires considerable effort.

Alexa Skills Kit (ASK)\footnote{\url{https://developer.amazon.com/alexa/alexa-skills-kit}}, Google Dialogflow\footnote{\url{https://cloud.google.com/dialogflow}}, and Voiceflow\footnote{\url{https://www.voiceflow.com/}} are representatives of the category of online platforms. ASK and Dialogflow offer an online application for the definition of NLU models (intents and slots/entities). For dialogue management, ASK introduced Alexa Conversations \citep{acharya-etal-2021-alexa} allowing the developer to specify the conversation flow using sample conversations. Dialogflow CX (a part of Dialogflow providing a new way of working with conversational agents) offers a visual view of the conversation agents to see the possible flows of the conversation. Voiceflow uses a visual dialogue editor as an approach to building a conversational application. Dialogue blocks can be connected via transitions to create a flow of blocks where each block represents a specific functionality such as keyword matching or API calls.

The comparison of the key aspects of the described libraries and platforms is shown in Table~\ref{tab:compare}.

\begin{table}[h!]
\scalebox{0.9}{
\begin{tabular}{l|llllll}
\setlength\tabcolsep{9pt}
                           & \rot{\textbf{Flowstorm}} & \rot{Rasa} & \rot{DeepPavlov} & \rot{Amazon ASK} & \rot{Google Dialogflow} & \rot{Voiceflow} \\ \hline
SaaS            &     \checkmark    &               &                     &   \checkmark        &    \checkmark                 &     \checkmark    \\ 
Open-source     &     \checkmark    &  \checkmark   &     \checkmark      &                     &                              &   \textit{p}            \\ 
Visual editor   &     \checkmark    &               &                     &                     &            \textit{p}        &    \textit{p}             \\ 
Custom code     &     \checkmark    &  \checkmark   &                     &                     &                              &                    \\ 
ML-based NLU    &     \checkmark    &  \checkmark   &    \checkmark       &   \checkmark        &     \checkmark              &    \checkmark      \\ 
Sub-dialogues  &     \checkmark    &               &                     &                     &                              &            \\ 
Sharing assets  &     \checkmark    &               &                     &                     &                              &   \checkmark      \\ 
Built-in NRG    &     \checkmark    &               &                     &                     &                              &                    \\ 
Analytic tools  &     \checkmark    &               &                     &    \checkmark       &          \checkmark          &    \checkmark      \\ 

\end{tabular}
}
\caption{Comparison of frameworks and online conversational platforms, \textit{p} means partial functionality.}
\label{tab:compare}
\end{table}

%

\section{Architecture}

The Flowstorm platform combines two sets of functions---design and analytics. The design functions allow users to create individual sub-dialogues, train entity models, write shared code, and combine the sub-dialogues into complex conversational applications. Each of the created assets can then be shared with the community. The analytic functions can be used to inspect the applications' traffic and show basic metrics such as session transcripts or visualizations of specific parameters.

The following sections describe a novel conversational system architecture. The novel aspect consists of a combination of manually created sub-dialogues (that can be easily combined into a complex dialogue structure) and neural generative models. First, we describe the blocks essential for creating an application, and then we describe each component used in runtime in detail.

\subsection{Conversational Application}
Conversational application is a set of dialogue trees we call sub-dialogues. Each sub-dialogue is focused on a small subset of a domain (e.g., in the movie domain, one sub-dialogue can be focused on a user's favorite movie). Each application has at least one sub-dialogue, but typically consists of several sub-dialogues. The sub-dialogue which is triggered when the application is launched is called ``main''. The other sub-dialogues are triggered when a specific situation occurs during the conversation. It depends on the specific design of the application, and it is described in detail in the following subsection.

Additionally, each application has configurable parameters such as language, Skimmer rules, or voice.

\subsection{Sub-dialogue}
\begin{figure}[h!]
    \centering
    \includegraphics[width=0.9\columnwidth]{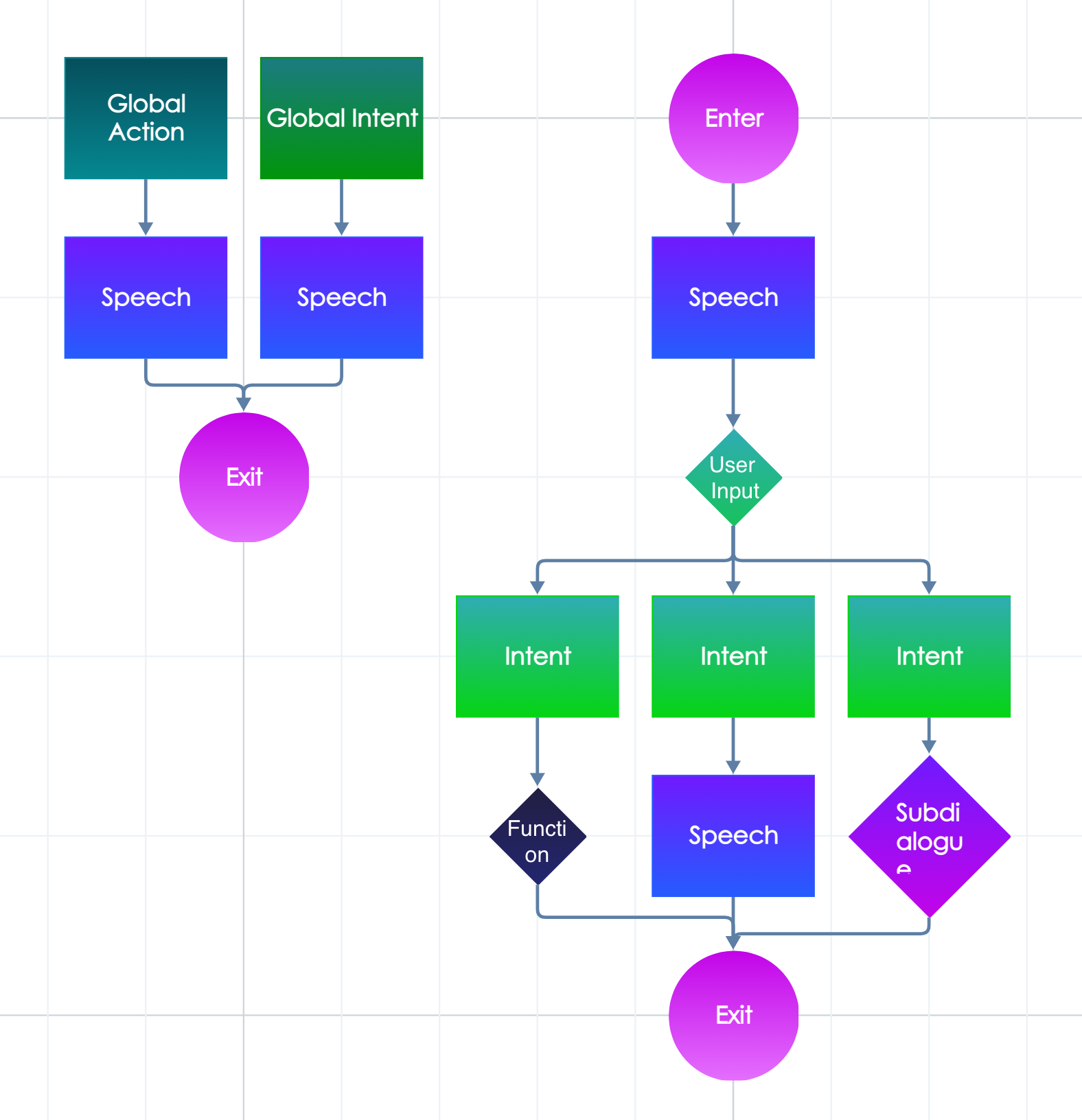}
    \caption{Nodes connected to the sub-dialogue structure. The names of the node types are shown in the corresponding boxes in the picture.}
    \label{fig:nodes}
\end{figure}

Sub-dialogue is a graph structure consisting of connected nodes representing the flow of the conversation. The key node types are described in the following list. The basic structure and node types are shown in Figure \ref{fig:nodes}.

    \textbf{Enter} -- Entry point of the sub-dialogue. Each sub-dialogue must have exactly one Enter.
    
    \textbf{Speech} -- Speech node contains one or more natural language responses that are presented to a user. If there are multiple responses, one is randomly selected. It can also contain slots that are eventually filled with variables based on the context. The responses can be customized using Kotlin-based DSL language.
    
    \textbf{User Input} -- Point in a conversation where the bot waits for the user utterance. It is typically connected to multiple intent nodes. Each User Input node has its own underlying intent recognition model trained separately. It has as many intent classes as there are intent nodes connected to it. The default behavior can be customized using Kotlin-based DSL language.
    
    \textbf{(Global) Intent} -- Intent node contains examples of the users' possible utterances which serve as training data for the intent recognition model. Each intent represents a single class in the intent recognition model. There are two types of intent nodes: contextual Intent and Global Intent. Contextual intents can be recognized only at a specific point of the conversation---when the conversation flow reaches the User Input node they are connected to. Global Intents, however, can be recognized at any point of the conversation.
    
    \textbf{Function} -- Function node contains code written in the Kotlin language \citep{kotlin}. The code can contain arbitrary logic (usually working with attributes --- see the attributes subsection). The function returns the transition to the next dialogue node. It is suitable for branching the dialogue based on the attribute values, API results, etc.
    
    \textbf{(Global) Action} -- Actions represent specific situations in the conversation flow. There are several predefined situations: \textit{Silence}, \textit{Error} and \textit{Out of domain}. \textit{Silence} action is triggered when no speech is recognized. \textit{Error} action is triggered when the logic written in the Function nodes fails (connection errors or bad design). \textit{Out of domain} action represents the utterance that does not belong to any of the intent classes. This will be described in detail in the following sections.
    
    \textbf{Sub-dialogue} -- Sub-dialogue node introduces the ability to reference another sub-dialogue. When the conversation flow reaches this node, it triggers the referenced sub-dialogue, processing its logic from the Enter node to the Exit.
    
    \textbf{Exit} -- Exit node represents the end of the sub-dialogue flow. If the sub-dialogue is the ``main'' dialogue, the session is ended. If it is a lower-level sub-dialogue, the conversation flow jumps to the parent sub-dialogue (i.e., the one where the current sub-dialogue was referenced).

\textbf{Init code} -- Each sub-dialogue has a declarative code part that typically serves as a place to define attributes (see below) and methods that can be called from the function nodes. As the code is written in the Kotlin language, the type control is done in the build phase, which lowers runtime errors.

\textbf{Attributes} -- There are four scopes of the attributes that can be defined in the Init code. The attribute stores values that can be used to modify the conversation flow, or they can be presented directly as part of a response. Each attribute must have its default value which is used to initialize the attribute based on its scope.

\begin{enumerate}
    \item \textit{Turn} -- the value is reset to default at the beginning of each turn.
    \item \textit{Session} -- the value is reset to default at the beginning of each session.
    \item \textit{User} -- the default value is used for each new user. Once the value of a user attribute is set, it is valid for all sessions of the user.
    \item \textit{Community} -- the default value is used for each community namespace. Once the value of a community attribute is set, it is valid for all users using the same community namespace.
\end{enumerate}

\section{System Components}\label{components}
The Flowstorm conversational platform consists of several components crucial for processing the user utterance and driving the dialogue. First, NLU gathers the information required by the DM. Afterward, DM traverses the corresponding sub-dialogue tree structure based on the NLU output.

\subsection{Natural Language Understanding}\label{NLU}
The NLU consists of the main components---entity recognition and intent recognition---that are commonly used in conversational systems. Additionally, the out-of-domain detection enhances the intent recognition, allowing the system to handle a situation that was not expected during dialogue design.

\subsubsection{Entity Recognition}
There are two modules used for Entity Recognition. Regex-based tool Duckling\footnote{\url{https://github.com/facebook/duckling}} and a sequence labeling neural network. The neural network uses the Bi-LSTM-CRF architecture \citep{huang2015bidirectional}. The architecture is simple enough to provide reasonable response time when run on CPU. Duckling is used to recognize such entities that have a specific structure and can be easily described using regular expressions (date, time, numbers, URLs, amount of money, etc.). Additionally, it parses (not only) the date and time values from natural language representation into a structured format. We enhanced the Duckling component with rules for numbers and time values in the Czech language.

During the process of dialogue creation, one can define custom entity types that can be recognized in the conversation. The entities are defined using example sentences where they can occur, e.g. \texttt{My favorite movie is [Matrix]\{movie\}}, meaning the word \texttt{Matrix} is an entity with type \texttt{movie}. These examples are used as training data for the sequence labeling model.

\subsubsection{Entity Masking}
The recognized entities are masked out in the user utterance prior to the intent recognition phase. This step allows the intent recognition model to focus only on the types of entities rather than the actual values. The intent recognition module receives, for example, the sentence \texttt{My favorite movie is \{movie\}} instead of \texttt{My favorite movie is Matrix}. Only the entity types included in the intent examples (in the intent nodes) are masked out.

\subsubsection{Intent Recognition}\label{sub:intent}
A key part of each sub-dialogue are the intent nodes. Each sub-dialogue may contain multiple global intents and multiple user input nodes with various numbers of local intents connected to it. Let the $ \mathcal{G} = (g_1, \ldots, g_n) $ be a set of global intents of the sub-dialogue $ S $ and let the $ \mathcal{U} = (u_1, \ldots, u_m) $ be the set of User Input nodes where each node $ u_i $ has its own set of local intents $ \mathcal{L}_i = (l_1, \ldots, l_{k_i}) $. There is one model trained for each $ u_i \in \mathcal{U} $ that classifies the user utterance into $ |\mathcal{L}_i| $ classes. Additionally, there is one extra model trained for global intents which has $ |\mathcal{G}| $ classes.

\begin{figure}[h!]
    \centering
    \includegraphics[width=\columnwidth]{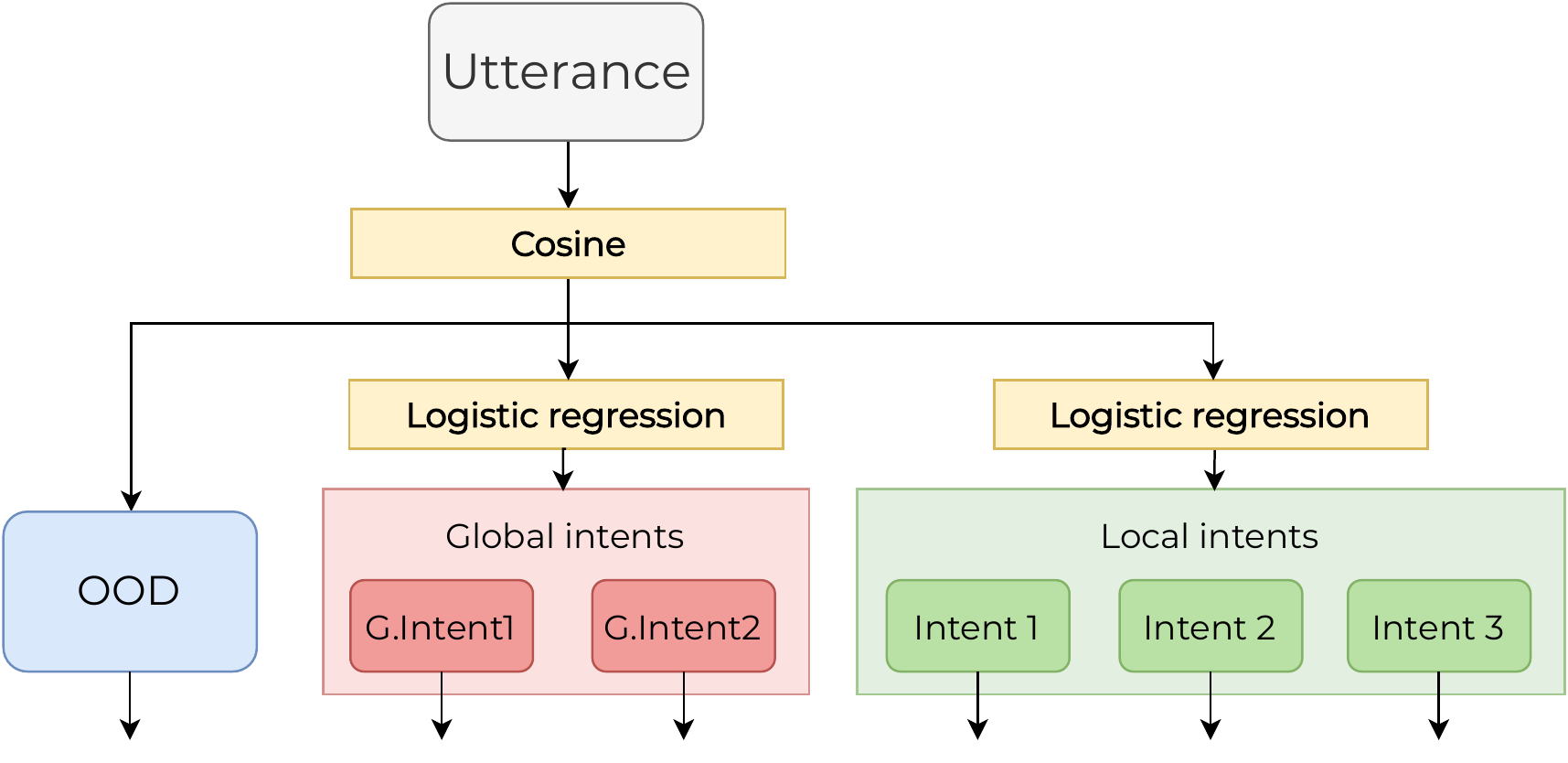}
    \caption{Classification algorithm for intent and OOD detection. The utterance is first classified by a cosine similarity into a class of local intents or a class of global intents. Next, the corresponding logistic regression makes the final intent classification. Additionally, cosine similarity can predict the OOD class if the similarity score falls below the threshold.}
    \label{fig:model}
\end{figure}

When the conversation flow reaches the User Input node, the system starts listening to the user (or waits for a text input). Then the actual intent recognition process consists of two steps:

First, the user utterance is embedded using the Universal Sentence Encoder \citep{cer2018universal} and the embedding is compared with the training examples from both global and local intent classes. This step only decides whether the final intent will be selected from local intents, global intents or whether the utterance is out of domain given the context of the current User Input node (see Subsection \textit{Out of Domain Detection} and Figure \ref{fig:model}). Note that not only global intents from the current sub-dialogue are considered but also the global intents from all parent sub-dialogues which are on the path from the current sub-dialogue to the main dialogue.
    
Second, the corresponding model (either global intent model or a model corresponding to the current User Input node) is selected. This model classifies the utterance into the final class.

The intent recognition experiments and a comparison with other NLU tools is described in the paper by \citet{lorenc2021benchmark}.

\subsubsection{Out of Domain Detection}\label{subsub:ood}
During each dialogue turn, the utterance is classified into one of the intent classes. However, the utterance does not have to correspond to either of the classes. We call such utterances out of domain (OOD). Whether the utterance is OOD or not depends purely on the sample utterances in the global intents and the corresponding local intent classes given in the current User Input node. The OOD is triggered using the confidences of the corresponding intent classes and the empirically estimated threshold. When the OOD is recognized, the dialogue flow continues with a local OOD action (if connected to the current User input node) or with a global OOD action.

\subsection{Skimmer}
The Skimmer is a component that extracts relevant data from the user utterance and stores it for later usage. It is done on a turn basis regardless of the dialogue context. An example of such information is: the information about the user having a brother mentioned ``by the way'' in the conversation (e.g., in a movie-related conversation, the user may mention \textit{``I went to the cinema with my brother yesterday.''}). The component skims through each utterance and saves the values into the attributes. The dialogue creator can then use the information in the conversation. The Skimmer is defined by a set of rules containing the following attributes:
\begin{itemize}
    \item \textit{Regular expression} - a set of patterns which must be contained in the utterance.
    \item \textit{Attribute name} - the name of the attribute where the value will be stored.
    \item \textit{Value} - the value stored in the attribute, typically \textit{true}, \textit{false}, or a matched group of the regular expression.
\end{itemize}
This component is inspired by a similar functionality used in the Alquist socialbot \citep{2021alquist}.

\subsection{Dialogue Management}
Dialogue management (DM) of a conversational application created in Flowstorm is divided into two stages: \textit{Sub-dialogue DM} and \textit{Dialogue selection}. The former stage operates on the graph structure of the corresponding sub-dialogue. The DM simply follows the directed edges beginning in the \textit{Start} node. There are two types of nodes that may have multiple outgoing edges: User Inputs and Functions. In the User Input nodes, the next edge is selected using Intent Recognition results, i.e., the edge leading to the most probable intent or global intent is selected. The Function nodes contain code that returns one of the connected edges.

\subsubsection{Dialogue selector}
The dialogue selector stage is triggered whenever a previous sub-dialogue is ended, and there is not a direct transition to a different one. The overall goal is to select a sub-dialogue that is related to the context of the conversation. The most suitable sub-dialogue is selected from a list of eligible sub-dialogues, which is defined by the developer who creates the dialogue application. Each sub-dialogue has several properties that are considered during the process of the dialogue selection.
\begin{itemize}
    \item \textit{Label} -- one or more labels can be assigned to each sub-dialogue.
    \item \textit{Entity} -- a sub-dialogue can use the information about an entity and work with the entity attributes. E.g., a sub-dialogue works with the information about a person, including the information about their birth date, occupation, etc. In that case, the sub-dialogue is tagged with the entity type Person.
    \item \textit{Starting conditions} is a custom logic that tells whether the corresponding sub-dialogue can be selected given the current context. E.g., in the case of the sub-dialogue discussing the user's favorite movie, it needs to check whether the information about the favorite movie is already stored. Additionally, most of the sub-dialogue can be triggered only once per session. Hence, the starting condition typically checks that as well.
\end{itemize}

The intuition behind the dialogue selector is to choose a dialogue whose starting condition is met, and its label and entity sets have the biggest overlap with the label and entity sets currently being discussed in the session. Formally, let $ \mathcal{D} $ be a set of eligible sub-dialogues, $ S $ be the current session. $ \operatorname{label}(d, S) $ returns the overlapping labels between the dialogue $ d $ and session $ S $, and, analogically,  $ \entity(d, S) $ returns overlapping entities, and $ \mathcal{C} $ is a set of starting conditions. Then the dialogue selector selects a sub-dialogue as follows:
\begin{align*}
    \argmax_{d \in \mathcal{D}} |\operatorname{label}(d, S) \cup \entity(d, S)| \\
    s.t.\hspace{0.2cm} \forall \cond_i \in \mathcal{C}: \cond_i(d, S)
\end{align*}

\subsection{Neural Response Generation}
All the principles described in the previous sections are suitable for creating high-quality content for various scenarios with maximum control over the conversation flow. A natural language conversation can, however, easily deviate from these scenarios. The Flowstorm platform comes with a novel approach using a combination of manually created content and generated content for this type of situation. During the dialogue creation process, the user may plug the generated response into any part of the sub-dialogue.

The platform currently uses a GPT-2-based \citep{gpt2} architecture for generating responses. The model takes a sequence of user utterances and bot responses as input. Additionally, the input contains the information about the dialogue act of the desired response. Since the model expects only natural language as the input, it is suitable for situations where the next response cannot be determined using only the tree-like dialogue structure.

\begin{figure}[h!]
    \centering
    \includegraphics[width=0.95\columnwidth]{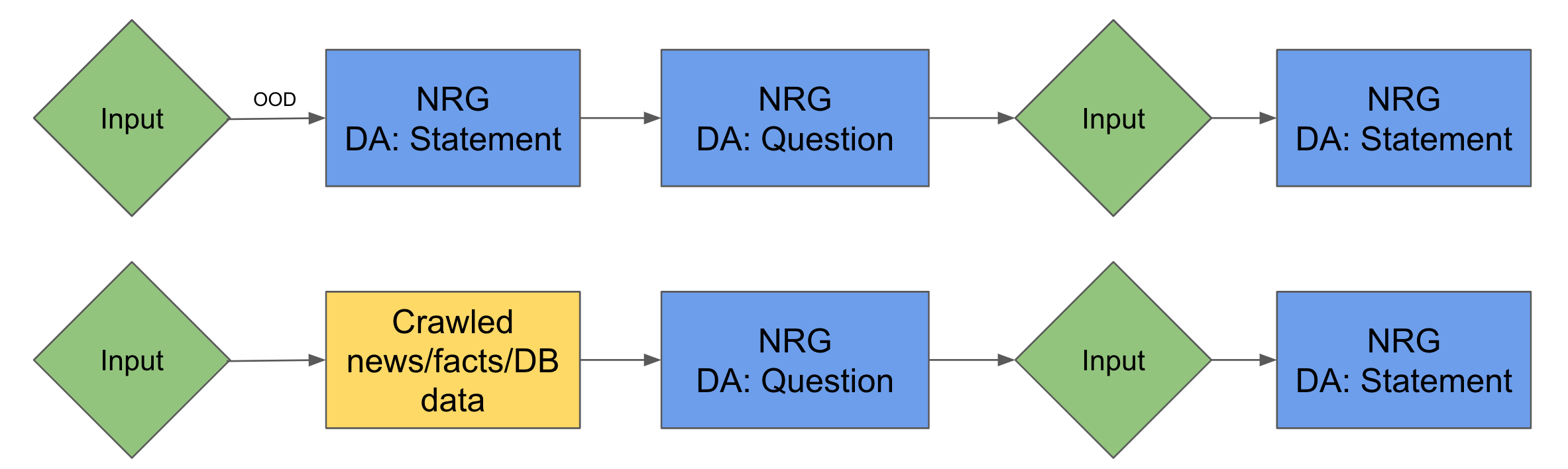}
    \caption{Two basic NRG usage scenarios. The upper part uses NRG response when OOD is detected. NRG generates a statement and a question, and then the system waits for the user's utterance. Afterward, another NRG statement is used, and the flow is returned to the dialogue graph. The lower part illustrates the scenario where the crawled text or information from a DB is used. A single NRG generated question follows it. The next turn starts with the NRG statement, and the flow is returned to the dialogue graph.}
    \label{fig:nrg}
\end{figure}

A typical use case for a generative model is to use it whenever an out-of-domain utterance is recognized. Since the dialogue designer cannot easily predict all out-of-domain variants and design a proper reaction, the generative approach is a suitable solution as it operates purely with the dialogue history.

In Figure \ref{fig:nrg}, we show two typical use cases of the generative model---generating response after an out-of-domain utterance is detected and asking additional questions based on free text (news article, fun-fact, etc.). 

\section{Analytics}
The set of analytic tools allows developers to inspect the traffic on their conversational applications. There are tools for inspecting session transcripts, community and user attributes, and creating metric visualizations. The \textit{Session transcripts} contains a list of the dialogue turns of each session with all the annotations and log messages. One can easily watch the NLU results, identify a weak part of the corresponding application, inspect the Automatic Speech Recognition (ASR) hypotheses, duration of each turn, and session attributes stored during the conversation. The community and user attributes can be viewed in individual tables to watch the values gathered during conversations.

\begin{figure}[h!]
    \centering
    \includegraphics[width=\columnwidth]{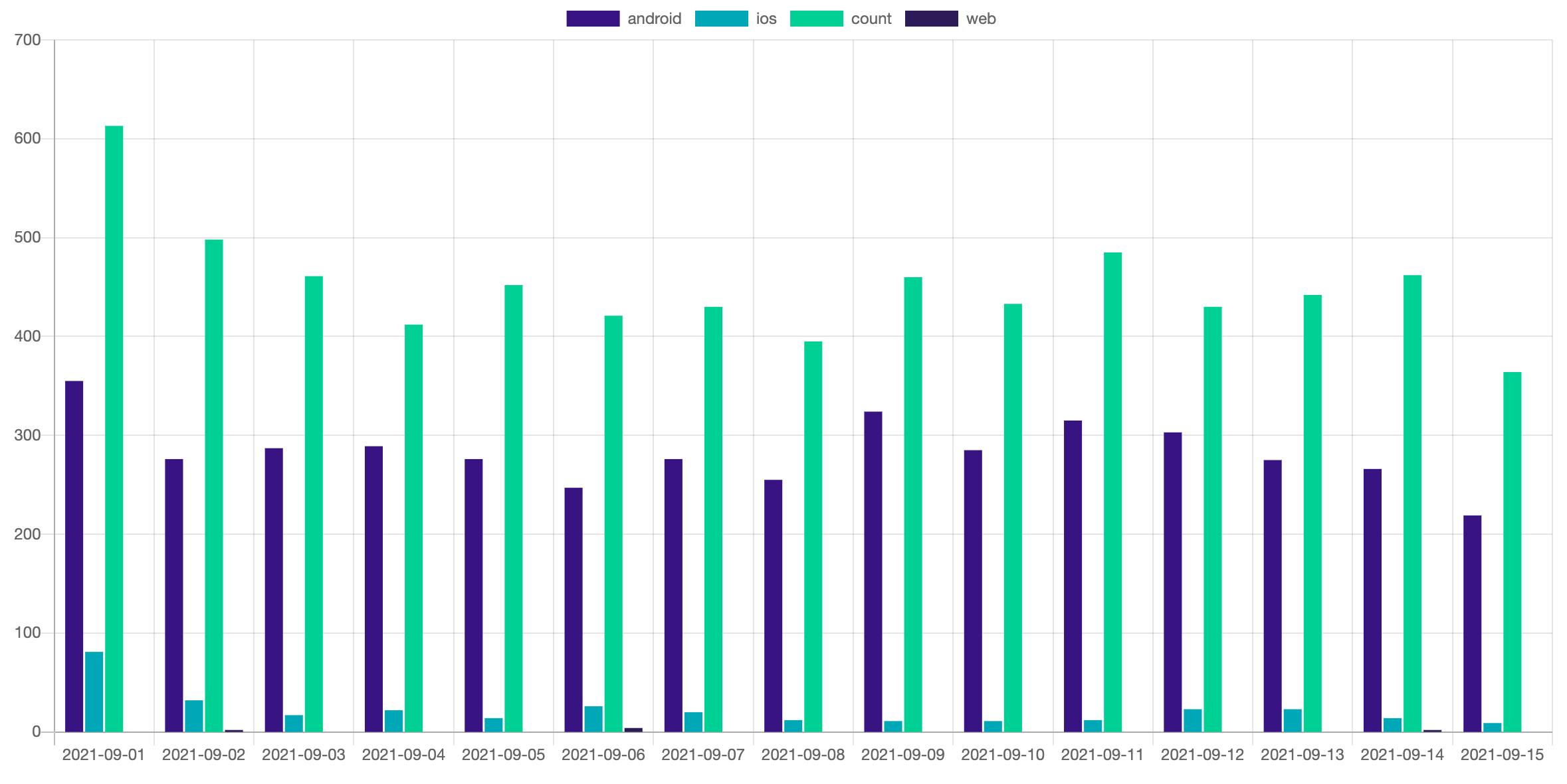}
    \caption{Example metric visualization. The columns show the count of sessions per client (Android, iOS, web and total).}
    \label{fig:analytics}
\end{figure}

The metric visualization tool allows creating plots of values defined as a metric. One can easily filter the specific values, time ranges, applications, users and specify the granularity of the visualization. An example visualization is shown in Figure \ref{fig:analytics}.

\section{Conclusion}
The open-source platform Flowstorm allows users to easily create conversational applications with low effort while allowing them to modify each part with custom functionality. Developers can visually design an application and extend it with custom code thanks to an easy-to-use interface. A library of shared assets contains common dialogue structures that can be used in various use cases. A novel hybrid architecture allows the system to handle unexpected utterances and make the conversation more robust. Analytic tools show the conversation transcripts, attributes, and metrics directly in the Flowstorm app. Based on the analytic data, the developer can quickly identify the weak parts of the application and fix them by modifying the dialogue structure or intent examples.

The Flowstorm platform is in active development, and we are primarily focusing on the generative part of the dialogue architecture to make it more suitable for various dialogue situations.
\bibliography{anthology,flowstorm}
\bibliographystyle{acl_natbib}

\end{document}